\documentclass[11pt]{article}

\usepackage[utf8]{inputenc} 
\usepackage[T1]{fontenc}    
\usepackage{hyperref}       
\usepackage{url}            
\usepackage{booktabs}       
\usepackage{amsfonts}       
\usepackage{nicefrac}       
\usepackage{microtype}      
\usepackage{times}
\usepackage{amsmath, amsfonts, amsthm, amssymb,epsfig}
\usepackage{hyperref}
\usepackage{amsfonts}
\usepackage{fullpage}
\usepackage{paralist}
\usepackage{bm}
\usepackage[ruled,vlined, linesnumbered]{algorithm2e}
\usepackage{url}
\usepackage{subfigure}
\usepackage{comment}
\usepackage{mathtools}
\usepackage{hyperref}
\usepackage{amsmath}

\usepackage{amsmath,amsfonts,amsthm,amssymb}
\usepackage{graphicx}
\usepackage{wrapfig}
\usepackage{lipsum}

\usepackage{authblk}

\def \x {\mathbf{x}}
\def \c {\tilde c}

\def \Z {\mathbf{X}}

\def \e {e}

\def \c {\mathbf{c}}

\def \e {\mathbf{e}}
\def \x {\mathbf{x}}

\def \R {\mathbb{R}}

\def \S {\mathcal{S}}

\def \a {\mathbf{a}}

\def \x {\mathbf{x}}
\def \c {\tilde c}

\def \d {\mathbf{d}}

\def \e {e}

\def \c {\mathbf{c}}
\def \A {\mathbf{A}}
\def \B {\mathbf{B}}
\def \X {\mathbf{X}}
\def \W {\mathbf{W}}
\def \U {\mathbf{U}}
\def \V {\mathbf{V}}

\def \e {\mathbf{e}}
\def \x {\mathbf{x}}

\def \R {\mathbb{R}}

\def \S {\mathcal{S}}

\def \W {\mathbf{W}}
\def \H {\mathbf{H}}

\def \N {\mathcal{N}}
\usepackage{multirow}

\usepackage{lipsum}
\usepackage{color}
\usepackage{enumitem}
\setlist[enumerate]{itemsep=0mm}

\title{Scalable Demand-Aware Recommendation}

\author{
  Jinfeng Yi\textsuperscript{1}\thanks{Now at Tencent AI Lab, Bellevue, WA, USA},\ \ Cho-Jui Hsieh\textsuperscript{2},\ \ Kush R. Varshney\textsuperscript{3},\ \ Lijun Zhang\textsuperscript{4},\ \ Yao Li\textsuperscript{2}\\
  \textsuperscript{1}AI Foundations Lab, IBM Thomas J. Watson Research Center, Yorktown Heights, NY, USA\\
  \textsuperscript{2}University of California, Davis, CA, USA\\
  \textsuperscript{3}IBM Research AI, Yorktown Heights, NY, USA\\
 \textsuperscript{4}National Key Laboratory for Novel Software Technology, Nanjing University, Nanjing, China\\
}

\begin{document}

\date{}

\maketitle

\begin{abstract}
Recommendation for e-commerce with a mix of durable and nondurable goods has characteristics that distinguish it from the well-studied media recommendation problem.  The demand for items is a combined effect of \emph{form utility} and \emph{time utility}, i.e., a product must both be intrinsically appealing to a consumer and the time must be right for purchase.  In particular for durable goods, time utility is a function of inter-purchase duration within product category because consumers are unlikely to purchase two items in the same category in close temporal succession.  Moreover, purchase data, in contrast to ratings data, is implicit with non-purchases not necessarily indicating dislike.  Together, these issues give rise to the positive-unlabeled demand-aware recommendation problem that we pose via joint low-rank tensor completion and product category inter-purchase duration vector estimation.  We further relax this problem and propose a highly scalable alternating minimization approach with which we can solve problems with millions of users and millions of items in a single thread. We also show superior prediction accuracies on multiple real-world data sets.
\end{abstract}

\section{Introduction}
\label{sec:intro}

E-commerce recommender systems aim to present items with high utility to the consumers \cite{LeeH2014}. Utility may be decomposed into \emph{form utility}: the item is desired as it is manifested, and \emph{time utility}: the item is desired at the given point in time \cite{Steiner1976}; recommender systems should take both types of utility into account.  Economists define items to be either \emph{durable goods} or \emph{nondurable goods} based on how long they are intended to last before being replaced \cite{Sexton2013}. A key characteristic of durable goods is the long duration of time between successive purchases within item categories whereas this duration for nondurable goods is much shorter, or even negligible.  Thus, durable and nondurable goods have differing time utility characteristics which lead to differing demand characteristics.

\begin{figure}
\centering
\includegraphics[width=0.8\columnwidth]{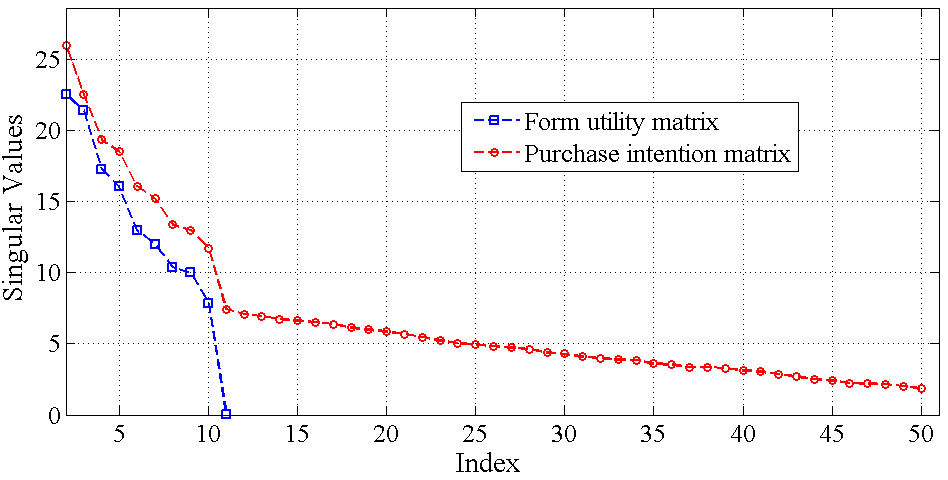}
\caption{\small{A toy example that illustrates the impact of time utility. It shows that although the form utility matrix is of low-rank (rank $10$), the purchase intention matrix is of full-rank (rank $50$).}}\label{fig:1}
\end{figure}

Although we have witnessed great success of collaborative filtering in media recommendation, we should be careful when expanding its application to general e-commerce recommendation involving both durable and nondurable goods due to the following reasons:
\leftmargini=8mm
\begin{enumerate}
\item Since media such as movies and music are nondurable goods, most users are quite receptive to buying or renting them in rapid succession. However, users only purchase durable goods when the time is right. For instance, most users will not buy televisions the day after they have already bought one.
    Therefore, recommending an item for which a user has no immediate demand can hurt user experience and waste an opportunity to drive sales.
\item A key assumption made by matrix factorization- and completion-based collaborative filtering algorithms is that the underlying rating matrix is of low-rank since only a few factors typically contribute to an individual's form utility~\cite{DBLP:journals/focm/CandesR09}. However, a user's demand is not only driven by form utility, but is the combined effect of both form utility and time utility. Hence, even if the underlying form utility matrix is of low-rank, the overall purchase intention matrix is likely to be of high-rank, and thus cannot be directly recovered by existing approaches. To see this, we construct a toy example with $50$ users and $100$ durable goods. Note that user $i$'s purchase intention of item $j$ is mediated by a time utility factor $h_{ij}$, which is a function of item $j$'s inter-purchase duration $d$ and the time gap $t$ of user $i$'s most recent purchase within the item $j$'s category. If $d$ and $t$ are Gaussian random variables, then the time utility $h_{ij}=\max(0,d-t)$ follows a rectified Gaussian distribution. Following the widely adopted low-rank assumption, we also assume that the form utility matrix $\Z\in \R^{50\times 100}$ is generated by $\U\V^\top$, where $\U\in \R^{50\times 10}$ and $\V\in \R^{100\times 10}$ are both Gaussian random matrices. Here we assume that $\U$, $\V$, and the time utility matrix $\H$ share the same mean (= $1$) and standard deviation (= $0.5$). Given the form utility $\X$ and time utility $\H$, the purchase intention matrix $\B\in \R^{50\times 100}$ is given by $\B=\Z-\H$. Figure~\ref{fig:1} shows the distributions of singular values for matrices $\Z$ and $\B$. It clearly shows that although the form utility matrix $\Z$ is of low-rank, the purchase intention matrix $\B$ is a full-rank matrix since all its singular values are greater than $0$. This simple example illustrates that considering users' demands can make the underlying matrix no longer of low-rank, thus violating the key assumption made by many collaborative filtering algorithms.
\end{enumerate}

An additional challenge faced by many real-world recommender systems is the \emph{one-sided sampling} of implicit feedback~\cite{DBLP:conf/icdm/HuKV08,DBLP:conf/uai/RendleFGS09}. Unlike the Netflix-like setting that provides both positive and negative feedback (high and low ratings), no negative feedback is available in many e-commerce systems. For example, a user might not purchase an item because she does not derive utility from it, or just because she was simply unaware of it or plans to buy it in the future. In this sense, the labeled training data only draws from the positive class, and the unlabeled data is a mixture of positive and negative samples, a problem usually referred to as \emph{positive-unlabeled  (PU) learning}~\cite{DBLP:conf/icml/HsiehND15}.
To address these issues, we study the problem of \emph{demand-aware recommendation}.  Given purchase triplets (user, item, time) and item categories, the objective is to make recommendations based on users' overall predicted combination of form utility and time utility.

We denote purchases by the sparse binary tensor ${\bm{\mathcal P}}$. To model implicit feedback, we assume that ${\bm{\mathcal P}}$ is obtained by thresholding an underlying real-valued utility tensor to a binary tensor ${\bm{\mathcal Y}}$ and then revealing a subset of ${\bm{\mathcal Y}}$'s positive entries.  The key to demand-aware recommendation is defining an appropriate utility measure for all (user, item, time) triplets. To this end, we quantify purchase intention as a combined effect of form utility and time utility. Specifically, we model a user's time utility for an item by comparing the time $t$ since her most recent purchase within the item's category and the item category's underlying inter-purchase duration $d$; the larger the value of $d-t$, the less likely she needs this item. In contrast, $d\le t$ may indicate that the item needs to be replaced, and she may be open to related recommendations. Therefore, the function $h=\max(0,\ d-t)$ may be employed to measure the time utility factor for a (user, item) pair. Then the purchase intention for a (user, item, time) triplet is given by $x-h$, where $x$ denotes the user's form utility. This observation allows us to cast demand-aware recommendation as the problem of learning users' form utility tensor ${\bm{\mathcal X}}$ and items' inter-purchase durations vector $\d$ given the binary tensor ${\bm{\mathcal P}}$.

Although the learning problem can be naturally formulated as a tensor nuclear norm minimization problem, the high computational cost significantly limits its application to large-scale recommendation problems. To address this limitation, we first relax the problem to a matrix optimization problem with a label-dependent loss. We note that the problem after relaxation is still non-trivial to solve since it is a highly non-smooth problem with nested hinge losses. More severely, the optimization problem involves
$mnl$ entries, where $m$, $n$, and $l$ are the number of users, items, and time slots, respectively. Thus a naive optimization algorithm will take at least $O(mnl)$ time, and is intractable for large-scale recommendation problems. To overcome this limitation, we develop an efficient alternating minimization algorithm and show that its time complexity is only approximately proportional to the number of nonzero elements in the purchase records tensor ${\bm{\mathcal P}}$. Since ${\bm{\mathcal P}}$ is usually very sparse, our algorithm is extremely efficient and can solve problems with millions of users and items.

Compared to existing recommender systems, our work has the following contributions and advantages: (i) to the best of our knowledge, this is the first work that makes demand-aware recommendation by considering inter-purchase durations for durable and nondurable goods; (ii) the proposed algorithm is able to simultaneously infer items' inter-purchase durations and users' real-time purchase intentions, which can help e-retailers make more informed decisions on inventory planning and marketing strategy; (iii) by effectively exploiting sparsity, the proposed algorithm is extremely efficient and able to handle large-scale recommendation problems.

\section{Related Work}
\label{sec:background}
Our contributions herein relate to three different areas of prior work: consumer modeling from a microeconomics and marketing perspective \cite{chatfield1973consumer}, time-aware recommender systems \cite{DBLP:journals/umuai/CamposDC14,SunPV2014,du2015time,DBLP:conf/ijcai/LiZLZXW11}, and PU learning \cite{DBLP:conf/icdm/LiuDLLY03,DBLP:conf/nips/PlessisNS14,DBLP:conf/icml/HsiehND15,hu2008collaborative,DBLP:conf/uai/RendleFGS09,baltrunas2009towards}.  The extensive consumer modeling literature is concerned with descriptive and analytical models of choice rather than prediction or recommendation, but nonetheless forms the basis for our modeling approach.  A variety of time-aware recommender systems have been proposed to exploit time information, but none of them explicitly consider the notion of time utility derived from inter-purchase durations in item categories.  Much of the PU learning literature is focused on the binary classification problem, e.g.\ \cite{DBLP:conf/icdm/LiuDLLY03,DBLP:conf/nips/PlessisNS14}, whereas we are in the collaborative filtering setting. For the papers that do examine collaborative filtering with PU learning or learning with implicit feedback~\cite{hu2008collaborative,DBLP:conf/uai/RendleFGS09,baltrunas2009towards,DBLP:conf/hcomp/YiJJJ13}, they mainly focus on media recommendation and overlook users' demands, thus are not suitable for durable goods recommendation.

Temporal aspects of the recommendation problem have been examined in a few ways: as part of the cold-start problem \cite{BobadillaOHB2012}, to capture dynamics in interests or ratings over time \cite{Koren2010}, and as part of the context in context-aware recommenders \cite{AdomaviciusT2011}.  However, the problem we address in this paper is different from all of those aspects, and in fact could be combined with the other aspects in future solutions.  To the best of our knowledge, there is no existing work that tries to take inter-purchase durations into account to better time recommendations as we do herein.

\section{Positive-Unlabeled Demand-Aware Recommendation}
\label{sec:formulation}
Throughout the paper, we use boldface Euler script letters, boldface capital letters, and boldface lower-case letters to denote tensors (e.g., ${\bm{\mathcal A}}$), matrices (e.g., $\A$) and vectors (e.g., $\a$), respectively. Scalars such as entries of tensors, matrices, and vectors are denoted by lowercase letters, e.g., $a$. In particular, the $(i,j,k)$ entry of a third-order tensor ${\bm{\mathcal A}}$ is denoted by $a_{ijk}$.

Given a set of $m$ users, $n$ items, and $l$ time slots, we construct a third-order binary tensor ${\bm{\mathcal P}}\in \{0,1\}^{m\times n \times l}$ to represent the purchase history. Specifically, entry $p_{ijk}=1$ indicates that user $i$ has purchased item $j$ in time slot $k$. We denote $\|\bm{\mathcal P}\|_0$ as the number of nonzero entries in tensor $\bm{\mathcal P}$. Since $\bm{\mathcal P}$ is usually very sparse, we have $\|\bm{\mathcal P}\|_0\ll mnl$. Also, we assume that the $n$ items belong to $r$ item categories, with items in each category sharing similar inter-purchase durations.\footnote{To meet this requirement, the granularity of categories should be properly selected. For instance, the category `Smart TV' is a better choice than the category `Electrical Equipment', since the latter category covers a broad range of goods with different durations.}
We use an $n$-dimensional vector $\c \in \{1,2,\ldots,r\}^n$ to represent the category membership of each item. Given ${\bm{\mathcal P}}$ and $\c$, we further generate a tensor ${\bm{\mathcal T}}\in \R^{m\times r \times l}$ where $t_{ic_jk}$ denotes the number of time slots between user $i$'s most recent purchase within item category $c_j$ until time $k$. If user $i$ has not purchased within item category $c_j$ until time $k$, $t_{ic_jk}$ is set to $+\infty$.

\subsection{Inferring Purchase Intentions from Users' Purchase Histories}
In this work, we formulate users' utility as a combined effect of form utility and time utility.  To this end, we use an underlying third-order tensor ${\bm{\mathcal X}}\in \R^{m\times n \times l}$ to quantify form utility. In addition, we employ a non-negative vector $\d\in \R_+^r$ to measure the underlying inter-purchase duration times of the $r$ item categories. It is understood that the inter-purchase durations for durable good categories are large, while for nondurable good categories are small, or even zero. In this study, we focus on items' inherent properties and assume that the inter-purchase durations are user-independent. The problem of learning personalized durations will be studied in our future work.

As discussed above, the demand is mediated by the time elapsed since the last purchase of an item in the same category.  Let $d_{c_j}$ be the inter-purchase duration time of item $j$'s category $c_j$, and let $t_{ic_jk}$ be the time gap of user $i$'s most recent purchase within item category $c_j$ until time $k$. Then if $d_{c_j}> t_{ic_jk}$, a previously purchased item in category $c_j$ continues to be useful, and thus user $i$'s utility from item $j$ is weak. Intuitively, the greater the value $d_{c_j} - t_{ic_jk}$, the weaker the utility. On the other hand, $d_{c_j} < t_{ic_jk}$ indicates that the item is nearing the end of its lifetime and the user may be open to recommendations in category $c_j$. We use a hinge loss $\max(0, d_{c_j} - t_{ic_jk})$ to model such time utility.  The overall utility can be obtained by comparing form utility and time utility. In more detail, we model a binary utility indicator tensor ${\bm{\mathcal Y}}\in \{0,1\}^{m\times n \times l}$ as being generated by the following thresholding process:
\begin{eqnarray}
y_{ijk}=\textbf{1}[x_{ijk}-\max(0, d_{c_j} - t_{ic_jk})>\tau]\label{eqn:B},
\end{eqnarray}
where $\textbf{1}(\cdot):\R \to \{0,1\}$ is the indicator function, and $\tau>0$ is a predefined threshold.

Note that the positive entries of ${\bm{\mathcal Y}}$ denote high purchase intentions, while the positive entries of ${\bm{\mathcal P}}$ denote actual purchases. Generally speaking, a purchase only happens when the utility is high, but a high utility does not necessarily lead to a purchase. This observation allows us to link the binary tensors ${\bm{\mathcal P}}$ and ${\bm{\mathcal Y}}$: ${\bm{\mathcal P}}$ is generated by a one-sided sampling process that only reveals a subset of ${\bm{\mathcal Y}}$'s positive entries.
Given this observation, we follow~\cite{DBLP:conf/icml/HsiehND15} and include a label-dependent loss~\cite{scott2012calibrated} trading the relative cost of positive and unlabeled samples:
\begin{eqnarray}
\mathcal{L}({\bm{\mathcal X}},{\bm{\mathcal P}}) \!= \eta \!\!\!\sum_{ijk:\ p_{ijk}=1
} \!\!\!\max[1-(x_{ijk}-\max(0, d_{c_j}-t_{i c_j k})), 0]^2
+(1-\eta)\!\!\!\!\!\!\sum_{ijk:\ p_{ijk}=0} \!\!\!\!\!l(x_{ijk},0),\nonumber
\end{eqnarray}
where $l(x,c)=(x-c)^2$ denotes the squared loss.

In addition, the form utility tensor ${\bm{\mathcal X}}$ should be of low-rank to capture temporal dynamics of users' interests, which are generally believed to be dictated by a small number of latent factors~\cite{DBLP:conf/pkdd/NaritaHTK11}.

By combining asymmetric sampling and the low-rank property together, we jointly recover the tensor ${\bm{\mathcal X}}$ and the inter-purchase duration vector $\d$ by solving the following tensor nuclear norm minimization (TNNM) problem:
\begin{eqnarray}
\min_{\bm{\mathcal X}\in \R^{m\times n\times l},\ \d\in\R_+^r}\!\!\!\!\!\!\!\!&\!\!\!\!\!\!\!\!\!\!\!\!\!\!\!\!\!\!\!\!\!\!&\eta \!\!\!\sum_{ijk:\ p_{ijk}=1
} \!\!\!\max[1-(x_{ijk}-\max(0, d_{c_j}-t_{i c_j k})), 0]^2\nonumber\\
&&+\ (1-\eta) \!\!\sum_{ijk:\ p_{ijk}=0}  \!\!\!\!x_{ijk}^{2} +\lambda\ \|\bm{\mathcal X}\|_*,\label{eqn:5}
\end{eqnarray}
where $\|{\bm{\mathcal X}}\|_*$ denotes the tensor nuclear norm, a convex combination of nuclear norms of ${\bm{\mathcal X}}$'s unfolded matrices~\cite{DBLP:journals/pami/LiuMWY13}. Given the learned $\hat {\bm{\mathcal X}}$ and $\hat \d$, the underlying binary tensor ${\bm{\mathcal Y}}$ can be recovered by (\ref{eqn:B}).

We note that although the TNNM problem (\ref{eqn:5}) can be solved by optimization techniques such as block coordinate descent~\cite{DBLP:journals/pami/LiuMWY13} and ADMM~\cite{gandy2011tensor}, they suffer from high computational cost since they need to be solved iteratively with multiple SVDs at each iteration. An alternative way to solve the problem is tensor factorization~\cite{DBLP:conf/nips/0002O14}. However, this also involves iterative singular vector estimation and thus not scalable enough. As a typical example, recovering a rank $20$ tensor of size $500\times 500\times 500$ takes the state-of-the-art tensor factorization algorithm TenALS~\footnote{\scriptsize{\url{http://web.engr.illinois.edu/~swoh/software/optspace/code.html}}} more than $20,000$ seconds on an Intel Xeon $2.40$ GHz processor with $32$ GB main memory. 

\subsection{A Scalable Relaxation}
In this subsection, we discuss how to significantly improve the scalability of the proposed demand-aware recommendation model. To this end, we assume that an individual's form utility does not change over time, an assumption widely-used in many collaborative filtering methods \cite{DBLP:conf/icml/SalakhutdinovM08a,DBLP:conf/hcomp/YiJJJ13}. Under this assumption, the tensor ${\bm{\mathcal X}}$ is a repeated copy of its frontal slice $\x_{::1}$, i.e.,
\begin{eqnarray}
{\bm{\mathcal X}} = \x_{::1}\circ\e,
\end{eqnarray}
where $\e$ is an $l$-dimensional all-one vector and the symbol $\circ$ represents the outer product operation. In this way, we can relax the problem of learning a third-order tensor ${\bm{\mathcal X}}$ to the problem of learning its frontal slice, which is a second-order tensor (matrix). For notational simplicity, we use a matrix $\Z$ to denote the frontal slice $\x_{::1}$, and use $x_{ij}$ to denote the entry $(i,j)$ of the matrix $\Z$.

Since ${\bm{\mathcal X}}$ is a low-rank tensor, its frontal slice $\Z$ should be of low-rank as well. Hence, the minimization problem \eqref{eqn:5} simplifies to:
\begin{eqnarray}
\min_{\Z\in\R^{m\times n}\atop \d\in \R^{r}} \!\!&&\!\!\!\!\eta\!\!\!\! \sum_{ijk:\ p_{ijk}=1} \max[1-(x_{ij}-\max(0, d_{c_j}-t_{i c_j k})), 0]^2\nonumber\\
&&+\ (1-\eta) \!\!\!\!\sum_{ijk:\ p_{ijk}=0}  \!\!\!\!x_{ij}^2 +\lambda\ \|\Z\|_*:=f(\Z, \d),\label{eqn:6}
\end{eqnarray}

where $\|\Z\|_*$ stands for the matrix nuclear norm, the convex surrogate of the matrix rank function. By relaxing the optimization problem \eqref{eqn:5} to the problem \eqref{eqn:6}, we recover a matrix instead of a tensor to infer users' purchase intentions.

\section{Optimization}
\label{sec:optimization}
Although the learning problem has been relaxed, optimizing \eqref{eqn:6} is still very challenging for two reasons: (i) the objective is highly non-smooth with nested hinge losses, and (ii) it contains $mnl$ terms: a naive optimization algorithm will take at least $O(mnl)$ time. 

To address these challenges, we adopt an alternating minimization scheme that iteratively fixes one of $\d$ and $\Z$ and minimizes with respect to the other. Specifically,
we apply an alternating minimization scheme to iteratively solve the following subproblems:
  \begin{align}
  \d & \leftarrow \arg\min_{\d} f(\Z, \d).  \label{eq:update_d}\\
    \Z &\leftarrow \arg\min_{\Z} f(\Z, \d)  \label{eq:update_Z}
  \end{align}
We note that both subproblems are non-trivial to solve because subproblem~\eqref{eq:update_Z} is a nuclear norm minimization problem, and both subproblems involve nested hinge losses.
In the following we discuss how to efficiently optimize subproblems~\eqref{eq:update_d} and~\eqref{eq:update_Z}:

\subsection{Update $\d$}
  Eq~\eqref{eq:update_d} can be written as
  \begin{equation*}
    \min_{\d} \sum_{ijk:\ p_{ijk}=1} \left\{ \max\bigg(1-(z_{ij}-\max(0, d_{c_j}-t_{ic_j k})),0\bigg)^2  \right\}:=g(\d):=\!\!\!\sum_{ijk:\ p_{ijk}=1} g_{ijk}(d_{c_j}).
  \end{equation*}
 We then analyze the value of each $g_{ijk}$ by comparing $d_{c_j}$ and $t_{i c_j k}$:
  \begin{enumerate}
    \item If $d_{c_j} \leq t_{i c_j k}$, we have
      \begin{equation*}
        g_{ijk}(d_{c_j})= \max(1-z_{ij}, 0)^2
      \end{equation*}
    \item If $d_{c_j} > t_{i c_j k}$, we have
    \begin{equation*}
     g_{ijk}(d_{c_j}) = \max(1-(z_{ij}-d_{c_j}+t_{ic_jk}),0)^2,
     \end{equation*}
    which can be further separated into two cases:
    \begin{equation*}
        g_{ijk}(d_{c_j}) = \begin{cases}
          1-(z_{ij}-d_{c_j}+t_{ic_j k}))^2, & \text{ if } d_{c_j} > z_{ij}+t_{i c_j k}-1\\
          0, & \text{ if } d_{c_j} \leq z_{ij}+t_{i c_j k}-1
        \end{cases}
      \end{equation*}

    \end{enumerate}
  Therefore, we have the following observations:
  \begin{enumerate}
    \item If $z_{ij}\leq 1$, we have
      \begin{equation*}
        g_{ijk}(d_{c_j}) = \begin{cases}
          \max(1-z_{i,j}, 0)^2, & \text{ if } d_{c_j} \leq t_{i c_j k} \\
          (1-(z_{ij}-d_{c_j}+t_{ic_jk}))^2, & \text{ if } d_{c_j} > t_{i c_j k}
        \end{cases}
      \end{equation*}
    \item If $z_{ij}> 1$, we have
      \begin{equation*}
        g_{ijk}(d_{c_j}) = \begin{cases}
          (1-(z_{ij}-d_{c_j}+t_{i c_j k}))^2, & \text{ if } d_{c_j} > t_{i c_j k} + z_{ij}-1\\
          0, & \text{ if } d_{c_j} \leq t_{i c_j k} + z_{ij} -1
        \end{cases}
      \end{equation*}
    \end{enumerate}

    This further implies
    \begin{equation*}
      g_{ijk}(d_{c_j}) = \begin{cases}
        \max(1-z_{ij},0)^2, &\text{ if } d_{c_j} \leq t_{i c_j k} + \max(z_{ij}-1,0) \\
        (1-(z_{ij}-d_{c_j}+t_{i c_j k}))^2, &\text{ if } d_{c_j} > t_{i c_j k} + \max(z_{ij}-1, 0)
      \end{cases}
    \end{equation*}
    For notational simplicity, we let $s_{ijk}=t_{i c_j k}+\max(z_{ij}-1, 0)$ for all triplets $(i,j,k)$ satisfying $\ p_{ijk}=1$.

\vspace{1.5mm}
    \noindent {\bf Algorithm. } For each category $\kappa$, we collect the set $Q=\{(i, j, k) \mid p_{ijk}=1 \text{ and } c_j=\kappa\}$ and calculate the corresponding $s_{ijk}$s.
    We then sort $s_{ijk}$s such that $(s_{i_1 j_1 k_1}) \leq \dots \leq s_{(i_{|Q|} j_{|Q|} k_{|Q|)}}$. For each interval $[s_{(i_q j_q k_q)}, s_{(i_{q+1} j_{q+1} k_{q+1}})]$,
    the function is
    \begin{equation*}
      g_{\kappa}(d) = \sum_{t=q+1}^{|Q|} \max(1-z_{i_t j_t}, 0)^2 + \sum_{t=1}^{q} (d+1 - z_{i_t j_t}-t_{i_t c_{j_t} k_t})^2
    \end{equation*}
    By letting
    \begin{align*}
      R_q &= \sum_{t=q+1}^{|Q|} \max(1-z_{i_t j_t},0)^2,\\
          F_q &= \sum_{t=1}^{q}(1-z_{i_t j_t}-t_{i_t c_{j_t} k_t}), \\
          W_q &= \sum_{t=1}^{q}(1-z_{i_t j_t}-t_{i_t c_{j_t}k_t})^2,
    \end{align*}
    we have
    \begin{align*}
      g_{\kappa}(d) &= qd^2 + 2F_q d + W_q  +   R_q \\
      &= q \bigg( d + \frac{F_q}{q} \bigg)^2 - \frac{F_q^2}{q} + W_q + R_q.
    \end{align*}
    Thus the optimal solution in the interval $[s_{(i_q j_q k_q)}, s_{(i_{q+1} j_{q+1} k_{q+1})}]$ is given by
    \begin{equation*}
      d^* = \max\left(s_{(i_q j_q k_q)},\ \min\big(s_{(i_{q+1} j_{q+1} k_{q+1})},\ -\frac{F_q}{q} \big) \right),
    \end{equation*}
    and the optimal function value is $g_r(d^*)$. By going through all the
    intervals from small to large, we can obtain the optimal solution for the
    whole function. We note that each time when $q\Rightarrow q+1$, the
    constants $R_q, F_q, W_q$ only change by one element. Thus the time
    complexity for going from $q\Rightarrow q+1$ is $O(1)$, and the whole
    procedure has time
    complexity $O(|Q|)$.

    In summary, we can solve the subproblem~\eqref{eq:update_d} by the
    following steps:
    \begin{enumerate}
    \item generate the set $U_\kappa=\{(i, j, k) \mid p_{ijk}=1 \text{ and } c_j=\kappa\}$ for each category $r$,
    \item sort each list (costing $O(|Q_\kappa|\log |Q_\kappa|)$ time),
    \item compute $R_0, F_0, W_0$ (costing $O(|Q_\kappa|)$ time), and then
    \item search for the optimal solution for each $q=1, 2, \cdots, |Q_\kappa|$ (costing $O(|Q_\kappa|)$ time).
    \end{enumerate}
    The above steps lead to an overall time complexity $O(\|{\bm{\mathcal P}}\|_0\log(\|{\bm{\mathcal P}}\|_0))$, where $\|{\bm{\mathcal P}}\|_0$ is the number of nonzero elements in tensor ${\bm{\mathcal P}}$. Therefore, we can efficiently update $\d$ since ${\bm{\mathcal P}}$ is a very sparse tensor with only a small number of nonzero elements.

  \subsection{Update $\Z$}
  By defining
   \begin{equation*}
        a_{i jk} = \begin{cases}
          1 + \max (0, d_{c_j}-t_{i c_j k}), & \text{ if } p_{ijk}=1\\
          0, & \text{\ otherwise}
        \end{cases}
      \end{equation*}
  the subproblem~\eqref{eq:update_Z} can be written as
  \begin{equation*}
    \min_{\Z\in\R^{m\times n}} h(\Z) + \lambda\|\Z\|_* \ \text{ where } h(\Z):=\bigg\{ \eta \sum_{ijk:\ p_{ijk}=1} \max(a_{ijk} - z_{ij},0)^2
    + (1-\eta) \!\!\!\sum_{ijk:\ p_{ijk}=0} z_{ij}^2 \bigg\}.
  \end{equation*}
  Since there are $O(mnl)$ terms in the objective function, a naive implementation will take $O(mnl)$ time, which is computationally inefficient when the data is large. To address this issue, We use proximal gradient descent to solve the problem. At each iteration, $\Z$ is updated by
  \begin{equation}
    \Z\leftarrow S_{\lambda}(\Z- \alpha\nabla h(\Z)),
  \end{equation}
  where $S_{\lambda}(\cdot)$ is the soft-thresholding operator for singular values \footnote{If $\X$ has the singular value decomposition
  $\X=\U\Sigma \V^T$, then $\S_{\lambda}(\X)=\U(\Sigma-\lambda I)_{+}\V^T$ where $a_+=\max(0,a)$. }.

   In order to efficiently compute the top singular vectors of $\Z-\alpha\nabla h(\Z)$, we rewrite it as
\begin{equation}
\Z - \alpha\nabla h(\Z) = [1 - 2(1-\eta)l]\ \Z + \left( 2(1-\eta) \sum_{ijk:\ p_{ijk}=1} z_{ij} - 2\eta \sum_{ijk:\ p_{ijk}=1} \max(a_{ijk}-z_{ij},0)\right).
\end{equation}

    \begin{algorithm}[t]
    \caption{Proximal Gradient Descent for Updating $\Z$}
  \label{alg:prox_grad}
  \SetKwInOut{Input}{Input}\SetKwInOut{Output}{Output}
  \Input{${\bm{\mathcal P}}$, $\Z^0$ (initialization), step size $\gamma$}
  \Output{A sequence of $\Z^t$ converges to the optimal solution}
  \For{$t=1,\dots,\text{maxiter}$} {
  $[\U, \Sigma, \V] = \text{rand\_svd}(\Z-\gamma \nabla h(\Z^t) )$ \\
  $\bar{\Sigma} = \max(\Sigma-\gamma\lambda,0)$ \\
  $k: \text{number of nonzeros in } \Sigma$ \\
  $\Z^{t+1} = \U(:, 1$:$k)\bar{\Sigma}(1$:$k, 1$:$k) \V(:, 1$:$k)^T$
  }
\end{algorithm}

Since $\Z$ is a low-rank matrix, $[1 - 2(1-\eta)l]\ \Z$ is also of low-rank. Besides, since ${\bm{\mathcal P}}$ is very sparse, the term
\[\left( 2(1-\eta) \sum_{ijk:\ p_{ijk}=1} z_{ij} - 2\eta \sum_{ijk:\ p_{ijk}=1} \max(a_{ijk}-z_{ij},0)\right)\]
is also sparse because it only involves the nonzero elements of ${\bm{\mathcal P}}$. In this case, when we multiply $(\Z-\alpha\nabla h(\Z))$ with a skinny $m$ by $k$ matrix, it can be computed in $O(nk^2+mk^2+\|{\bm{\mathcal P}}\|_0k)$ time.

As shown in~\cite{CJH13c}, each iteration of proximal gradient descent for
nuclear norm minimization only requires a fixed number of
iterations before convergence, thus the time complexity to update $\Z$ is
$O(nk^2T+mk^2T+\|{\bm{\mathcal P}}\|_0kT)$, where $T$ is the number of iterations.

  \subsection{Overall Algorithm}

  Combining the two subproblems together, the time complexity of each iteration of the proposed algorithm is:
\begin{equation*}
  O( \|{\bm{\mathcal P}}\|_0\log(\|{\bm{\mathcal P}}\|_0) + nk^2T +mk^2 T + \|{\bm{\mathcal P}}\|_0 k T).
\end{equation*}

\noindent\textbf{Remark:} Since each user should make at least one purchase and each item should be purchased at least once to be included in ${\bm{\mathcal P}}$, $n$ and $m$ are smaller than $\|{\bm{\mathcal P}}\|_0$.
  Also, since $k$ and $T$ are usually very small, the time complexity to solve problem~\eqref{eqn:6} is dominated by the term $\|{\bm{\mathcal P}}\|_0$, which is a significant improvement over the naive approach with $O(mnl)$ complexity.

Since our problem has only two blocks $\d,\ \Z$ and each subproblem is convex, our optimization algorithm is guaranteed to converge to a stationary point~\cite{LG00a}. Indeed, it converges very fast in practice. As a concrete example, it takes only $10$ iterations to optimize a problem with $1$ million users, $1$ million items, and more than $166$ million purchase records.

\section{Experiments}
\label{sec:empirical}

\subsection{Experiment with Synthesized Data}
We first conduct experiments with simulated data to verify that the proposed demand-aware recommendation algorithm is computationally efficient and robust to noise.
To this end, we first construct a low-rank matrix $\Z=\W\H^T$, where $\W\in\R^{m\times 10}$ and $\H\in\R^{n\times 10}$ are random Gaussian matrices with entries drawn from $\N(1,0.5)$, and then normalize $\Z$ to the range of $[0, 1]$. We randomly assign all the $n$ items to $r$ categories, with their inter-purchase durations $\d$ equaling
$[10,20,\ldots,10r]$. We then construct the high purchase intension set $\Omega=\{(i,j,k)\mid t_{ic_jk}\geq d_{c_j} \text{ and } x_{ij} \geq 0.5\}$, and sample a subset of its entries as the observed purchase records. We let $n=m$ and vary them in the range $\{10,000, 20,000, 30,000, 40,000\}$. We also vary $r$ in the range $\{10, 20, \cdots, 100\}$. Given the learned durations $\d^*$, we use ${\|\d-\d^*\|_2}/{\|\d\|_2}$ to measure the prediction errors.

  \begin{figure*}[tb]
  \centering
  \begin{tabular}{ccc}
    \subfigure[\scriptsize{Error vs Number of users/items}]{
    \label{fig:size}
    \includegraphics[width=0.302\linewidth,height=0.21\linewidth]{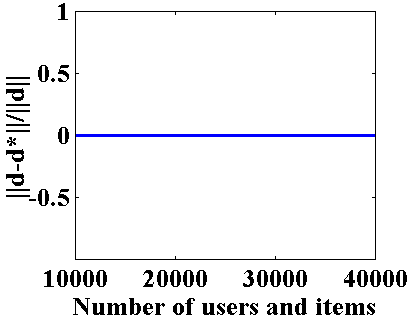}
    } &
    \subfigure[\scriptsize{Error vs Number of categories}]{
    \label{fig:categories}
    \includegraphics[width=0.294\linewidth,height=0.21\linewidth]{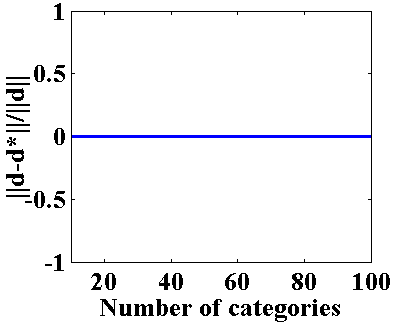}
    } &
    \subfigure[\scriptsize{Error vs Noise levels}]{
    \label{fig:noise}
    \includegraphics[width=0.30\linewidth,height=0.21\linewidth]{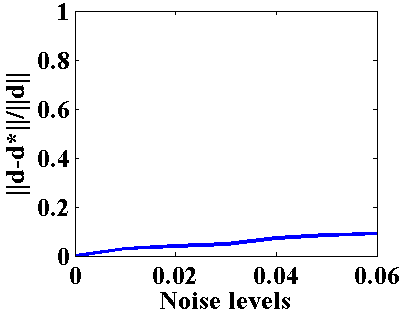}
    }
  \end{tabular}
  \caption{Prediction errors ${\|\d-\d^*\|_2}/{\|\d\|_2}$ as a function of number of users, items, categories, and noise levels on synthetic data sets \label{fig:synthetic}
}
\end{figure*}
 \begin{table*}[t]
   \caption{CPU time for solving problem \eqref{eqn:6} with different number of purchase records\label{tab:scalability}}
    \centering
  \begin{tabular}{|c|c|c|c|c|c|}
      \hline
      $m$ (\# users)   & $n$ (\# items) & $l$ (\# time slots) &$\|{\bm{\mathcal P}}\|_0$&\ $k$\ &  CPU Time (in seconds)  \\
      \hline
   1,000,000 & 1,000,000 & 1,000 & 693,826  & 10 & 250 \\
   \hline
   1,000,000 & 1,000,000 & 1,000 & 2,781,040  & 10 & 311  \\
   \hline
   1,000,000 & 1,000,000 & 1,000 & 11,112,400  & 10 & 595  \\
   \hline
   1,000,000 & 1,000,000 & 1,000 & 43,106,100  & 10 & 1,791  \\
   \hline
   1,000,000 & 1,000,000 & 1,000 & 166,478,000  & 10 & 6,496  \\
   \hline
  \end{tabular}
\end{table*}

\vspace{0.8mm}
\noindent \textbf{Accuracy}\ \ Figure~\ref{fig:size} and \ref{fig:categories} clearly show that the proposed algorithm can \emph{perfectly} recover the underlying inter-purchase durations with varied numbers of users, items, and categories. To further evaluate the robustness of the proposed algorithm, we randomly flip some entries in tensor ${\bm{\mathcal P}}$ from $0$ to $1$ to simulate the rare cases of purchasing two items in the same category in close temporal succession. Figure~\ref{fig:noise} shows that when the ratios of noisy entries are not large, the predicted durations $\hat \d$ are close enough to the true durations, thus verifying the robustness of the proposed algorithm.

\vspace{0.8mm}
\noindent \textbf{Scalability}\ \ To verify the scalability of the proposed algorithm, we fix the numbers of users and items to be $1$ million, the number of time slots to be $1000$, and vary the number of purchase records (i.e., $\|{\bm{\mathcal P}}\|_0$). Table~\ref{tab:scalability} summarizes the running time of solving problem \eqref{eqn:6} on a computer with $32$ GB main memory using a single thread. We observe that the proposed algorithm is extremely efficient, e.g., even with $1$ million users, $1$ million items, and more than $166$ million purchase records, the running time of the proposed algorithm is less than $2$ hours.

\subsection{Experiment with Real-World Data}
In the real-world experiments, we evaluate the proposed demand-aware recommendation algorithm by comparing it with the six state-of the-art recommendation methods: (a) \textbf{M$^3$F}, maximum-margin matrix factorization \cite{DBLP:conf/icml/RennieS05}, (b) \textbf{PMF}, probabilistic matrix factorization \cite{DBLP:conf/icml/SalakhutdinovM08a}, (c) \textbf{WR-MF}, weighted regularized matrix factorization \cite{hu2008collaborative}, (d) \textbf{CP-APR}, Candecomp-Parafac alternating Poisson regression \cite{DBLP:journals/siammax/ChiK12}, (e) \textbf{Rubik}, knowledge-guided tensor factorization and completion method \cite{DBLP:conf/kdd/WangCGDKCMS15}, and (f) \textbf{BPTF}, Bayesian probabilistic tensor factorization \cite{DBLP:conf/sdm/XiongCHSC10}. Among them, M$^3$F and PMF are widely-used static collaborative filtering algorithms. We include these two algorithms as baselines to justify whether traditional collaborative filtering algorithms are suitable for general e-commerce recommendation involving both durable and nondurable goods. Since they require explicit ratings as inputs, we follow~\cite{baltrunas2009towards} to generate numerical ratings based on the frequencies of (user, item) consumption pairs. WR-MF is essentially the positive-unlabeled version of PMF and has shown to be very effective in modeling implicit feedback data. All the other three baselines, i.e., CP-APR, Rubik, and BPTF, are tensor-based methods that can consider time utility when making recommendations. We refer to the proposed recommendation algorithm as \textbf{Demand-Aware Recommender for One-Sided Sampling}, or \textbf{DAROSS} for short.

Our testbeds are two real-world data sets \emph{Tmall}\footnote{\url{http://ijcai-15.org/index.php/repeat-buyers-prediction-competition}} and \emph{Amazon Review}\footnote{\url{http://jmcauley.ucsd.edu/data/amazon/}}. Since some of the baseline algorithms are not scalable enough, we first conduct experiments on their subsets and then on the full set of \emph{Amazon Review}. In order to generate the subsets, we randomly sample $80$ item categories for \emph{Tmall data set} and select the users who have purchased at least $3$ items within these categories, leading to the purchase records of $377$ users and $572$ items. For \emph{Amazon Review data set}, we randomly select $300$ users who have provided reviews to at least $5$ item categories on Amazon.com. This leads to a total of $5,111$ items belonging to $11$ categories. Time information for both data sets is provided in days, and we have $177$ and $749$ time slots for \emph{Tmall} and \emph{Amazon Review} subsets, respectively. The full \emph{Amazon Review} data set is significantly larger than its subset. After removing duplicate items, it contains more than $72$ million product reviews from $19.8$ million users and $7.7$ million items that belong to $24$ item categories. The collected reviews span a long range of time: from May 1996 to July 2014, which leads to $6,639$ time slots in total. Comparing to its subset, the full set is a much more challenging data set both due to its much larger size and much lower sampling rate, i.e., many reviewers only provided a few reviews, and many items were only reviewed a small number of times.

 \begin{figure*}[tb]
  \centering
  \begin{tabular}{ccc}
    \subfigure[Category Prediction]{
    \includegraphics[width=0.461\linewidth,height=0.265\linewidth]{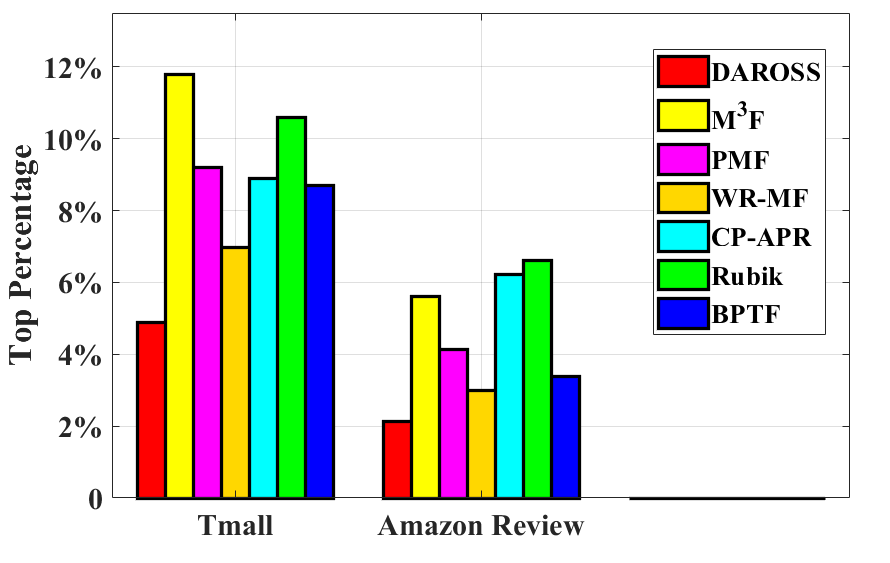}
    } &
    \subfigure[Purchase Time Prediction]{
    \includegraphics[width=0.461\linewidth,height=0.265\linewidth]{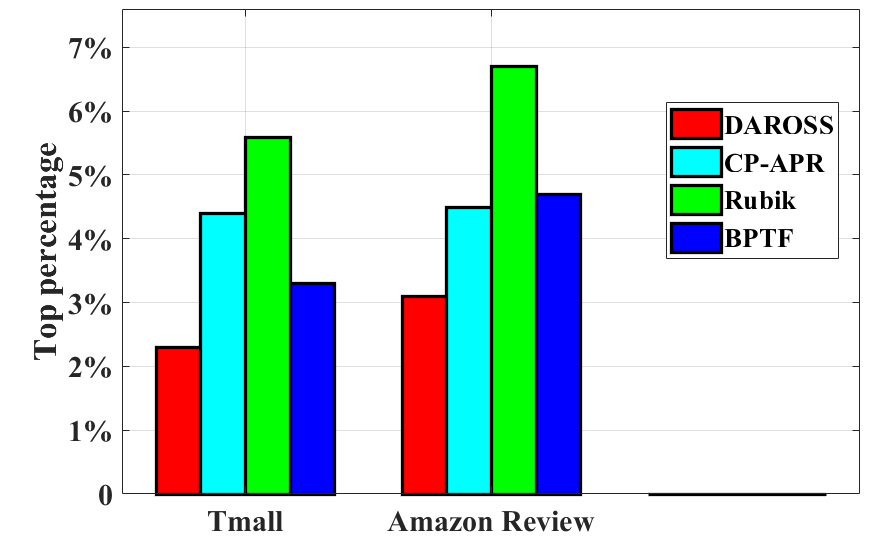}
    }
  \end{tabular}
  \caption{Prediction performance on real-world data sets \emph{Tmall} and \emph{Amazon Review} subsets\label{fig:real-world}
}
\end{figure*}
 \begin{table*}[tb]
   \caption{Estimated inter-reviewing durations for Amazon Review subset\label{tab:d}}
    \centering
  \begin{tabular}{|l|c|c|c|c|c|c|c|c|c|c|c|}
      \hline
    \multirow{2}{*}{\!\!\scriptsize{Categories}\!\!} &\!\!\scriptsize{Instant}\!\!  & \!\!\!\!\scriptsize{Apps for}\!\!\!\!& \!\!\scriptsize{Automotive}\!\! & \!\!\scriptsize{Baby}\!\! & \!\!\scriptsize{Beauty}\!\!\! &\!\!\scriptsize{Digital}\!\! & \!\!\scriptsize{Grocery }\!\! & \!\!\scriptsize{Musical}\!\! & \!\!\scriptsize{Office}\!\! & \scriptsize{Patio ...} & \!\!\scriptsize{Pet}\!\!\\
    &\scriptsize{Video}  & \scriptsize{Android} & & & &\scriptsize{Music} & \!\!\scriptsize{... Food}\!\! & \!\scriptsize{Instruments}\! & \scriptsize{Products} & \!\!\scriptsize{Garden}\!\! & \!\scriptsize{Supplies}\! \\
      \hline
   \footnotesize{$\d$}&\small{0}&\small{0}&\small{326}&\small{0}&\small{0}&\small{158}&\small{0}&\small{38}&\small{94}&\small{271}&\small{40}
   \\
   \hline
  \end{tabular}
\end{table*}

For each user, we randomly sample $90\%$ of her purchase records as the training data, and use the remaining $10\%$ as the test data.
For each purchase record ($u$, $i$, $t$) in the test set, we evaluate all the algorithms on two tasks: (i) \emph{category prediction}, and (ii) \emph{purchase time prediction}. In the first task, we record the highest ranking of items that are within item $i$'s category among all items at time $t$. Since a purchase record ($u$, $i$, $t$) may suggest that in time slot $t$, user $u$ needed an item that share similar functionalities with item $i$, \emph{category prediction} essentially checks whether the recommendation algorithms recognize this need.
In the second task, we record the number of slots between the true purchase time $t$ and its nearest predicted purchase time within item $i$'s category.
Ideally, good recommendations should have both small category rankings and small time errors. Thus we adopt the average top percentages, i.e., (\emph{average category ranking) / $n \times 100\%$} and (\emph{average time error) / $l \times 100\%$}, as the evaluation metrics of category and purchase time prediction tasks, respectively. The algorithms M$^3$F, PMF, and WR-MF are excluded from the purchase time prediction task since they are static models that do not consider time information.

Figure~\ref{fig:real-world} displays the predictive performance of the seven recommendation algorithms on \emph{Tmall} and \emph{Amazon Review} subsets. As expected, M$^3$F and PMF fail to deliver strong performance since they neither take into account users' demands, nor consider the positive-unlabeled nature of the data. This is verified by the performance of WR-MF: it significantly outperforms M$^3$F and PMF by considering the PU issue and obtains the second-best item prediction accuracy on both data sets (while being unable to provide a purchase time prediction). By taking into account both issues, our proposed algorithm DAROSS yields the best performance for both data sets and both tasks. Table~\ref{tab:d} reports the inter-reviewing durations of \emph{Amazon Review} subset estimated by our algorithm. Although they may not perfectly reflect the true inter-purchase durations, the estimated durations clearly distinguish between durable good categories, e.g., \emph{automotive}, \emph{musical instruments}, and non-durable good categories, e.g., \emph{instant video}, \emph{apps}, and \emph{food}.
Indeed, the learned inter-purchase durations can also play an important role in applications more advanced than recommender systems, such as inventory management, operations management, and sales/marketing mechanisms. We do not report the estimated durations of \emph{Tmall} herein since the item categories are anonymized in the data set.

Finally, we conduct experiments on the full \emph{Amazon Review} data set. In this study, we replace \emph{category prediction} with a more strict evaluation metric \emph{item prediction}~\cite{du2015time}, which indicates the predicted ranking of item $i$ among all items at time $t$ for each purchase record ($u$, $i$, $t$) in the test set. Since most of our baseline algorithms fail to handle such a large data set, we only obtain the predictive performance of three algorithms: DAROSS, WR-MF, and PMF. Note that for such a large data set, prediction time instead of training time becomes the bottleneck: to evaluate average item rankings, we need to compute the scores of all the 7.7 million items, thus is computationally inefficient. Therefore, we only sample a subset of items for each user and estimate the rankings of her purchased items. Using this evaluation method, the average item ranking percentages for DAROSS, WR-MF and PMF are $16.7\%$, $27.3\%$, and $38.4\%$, respectively. In addition to superior performance, it only takes our algorithm $10$ iterations and 1 hour to converge to a good solution. Since WR-MF and PMF are both static models, our algorithm is the only approach evaluated here that considers time utility while being scalable enough to handle the full \emph{Amazon Review} data set. Note that this data set has more users, items, and time slots but fewer purchase records than our largest synthesized data set, and the running time of the former data set is lower than the latter one. This clearly verifies that the time complexity of our algorithm is dominated by the number of purchase records instead of the tensor size.
Interestingly, we found that some inter-reviewing durations estimated from the full \emph{Amazon Review} data set are much smaller than the durations reported in Table~\ref{tab:d}. This is because the estimated durations tend to be close to the minimum reviewing/purchasing gap within each category, thus may be affected by outliers who review/purchase durable goods in close temporal succession. The problem of improving the algorithm robustness will be studied in our future work. On the other hand, this result verifies the effectiveness of the PU formulation -- even if the durations are underestimated, our algorithm still outperforms the competitors by a considerable margin. As a final note, we want to point out that \emph{Tmall} and \emph{Amazon Review} may not take full advantage of the proposed algorithm, since (i) their categories are relatively coarse and may contain multiple sub-categories with different durations, and (ii) the time stamps of \emph{Amazon Review} reflect the review time instead of purchase time, and inter-reviewing durations could be different from inter-purchase durations.
By choosing a purchase history data set with a more proper category granularity, we expect to achieve more accurate duration estimations and also better recommendation performance.

\section{Conclusion}\label{sec:conclusion}
In this paper, we examine the problem of demand-aware recommendation in settings when inter-purchase duration within item categories affects users' purchase intention in combination with intrinsic properties of the items themselves. We formulate it as a tensor nuclear norm minimization problem that seeks to jointly learn the form utility tensor and a vector of inter-purchase durations, and propose a scalable optimization algorithm with a tractable time complexity.
Our empirical studies show that the proposed approach can yield perfect recovery of duration vectors in noiseless settings; it is robust to noise and scalable as analyzed theoretically. On two real-world data sets, \emph{Tmall} and \emph{Amazon Review}, we show that our algorithm outperforms six state-of-the-art recommendation algorithms on the tasks of category, item, and purchase time predictions.

\bibliographystyle{plain}
\bibliography{IEEEabrv,durable,Sigir_Rank}

\begin{thebibliography}{10}

\bibitem{AdomaviciusT2011}
Gediminas Adomavicius and Alexander Tuzhilin.
\newblock Context-aware recommender systems.
\newblock In {\em Recommender Systems Handbook}, pages 217--253. Springer, New
  York, NY, 2011.

\bibitem{baltrunas2009towards}
Linas Baltrunas and Xavier Amatriain.
\newblock Towards time-dependant recommendation based on implicit feedback.
\newblock In {\em Workshop on context-aware recommender systems}, 2009.

\bibitem{BobadillaOHB2012}
Jes{\'u}s Bobadilla, Fernando Ortega, Antonio Hernando, and Jes{\'u}s Bernal.
\newblock A collaborative filtering approach to mitigate the new user cold
  start problem.
\newblock {\em Knowl.-Based Syst.}, 26:225--238, February 2012.

\bibitem{DBLP:journals/umuai/CamposDC14}
Pedro~G. Campos, Fernando D{\'{\i}}ez, and Iv{\'{a}}n Cantador.
\newblock Time-aware recommender systems: a comprehensive survey and analysis
  of existing evaluation protocols.
\newblock {\em User Model. User-Adapt. Interact.}, 24(1-2):67--119, 2014.

\bibitem{DBLP:journals/focm/CandesR09}
Emmanuel~J. Cand{\`{e}}s and Benjamin Recht.
\newblock Exact matrix completion via convex optimization.
\newblock {\em Foundations of Computational Mathematics}, 9(6):717--772, 2009.

\bibitem{chatfield1973consumer}
Christopher Chatfield and Gerald~J Goodhardt.
\newblock A consumer purchasing model with erlang inter-purchase times.
\newblock {\em Journal of the American Statistical Association},
  68(344):828--835, 1973.

\bibitem{DBLP:journals/siammax/ChiK12}
Eric~C. Chi and Tamara~G. Kolda.
\newblock On tensors, sparsity, and nonnegative factorizations.
\newblock {\em SIAM Journal on Matrix Analysis and Applications},
  33(4):1272--1299, 2012.

\bibitem{du2015time}
Nan Du, Yichen Wang, Niao He, Jimeng Sun, and Le~Song.
\newblock Time-sensitive recommendation from recurrent user activities.
\newblock In {\em NIPS}, pages 3474--3482, 2015.

\bibitem{DBLP:conf/nips/PlessisNS14}
Marthinus~Christoffel du~Plessis, Gang Niu, and Masashi Sugiyama.
\newblock Analysis of learning from positive and unlabeled data.
\newblock In {\em NIPS}, pages 703--711, 2014.

\bibitem{gandy2011tensor}
Silvia Gandy, Benjamin Recht, and Isao Yamada.
\newblock Tensor completion and low-n-rank tensor recovery via convex
  optimization.
\newblock {\em Inverse Problems}, 27(2):025010, 2011.

\bibitem{LG00a}
L.~Grippo and M.~Sciandrone.
\newblock On the convergence of the block nonlinear {G}auss-{S}eidel method
  under convex constraints.
\newblock {\em Operations Research Letters}, 26:127--136, 2000.

\bibitem{CJH13c}
C.-J. Hsieh and P.~A. Olsen.
\newblock Nuclear norm minimization via active subspace selection.
\newblock In {\em ICML}, 2014.

\bibitem{DBLP:conf/icml/HsiehND15}
Cho{-}Jui Hsieh, Nagarajan Natarajan, and Inderjit~S. Dhillon.
\newblock {PU} learning for matrix completion.
\newblock In {\em ICML}, pages 2445--2453, 2015.

\bibitem{hu2008collaborative}
Y.~Hu, Y.~Koren, and C.~Volinsky.
\newblock Collaborative filtering for implicit feedback datasets.
\newblock In {\em ICDM}, pages 263--272. IEEE, 2008.

\bibitem{DBLP:conf/icdm/HuKV08}
Yifan Hu, Yehuda Koren, and Chris Volinsky.
\newblock Collaborative filtering for implicit feedback datasets.
\newblock In {\em ICDM}, pages 263--272, 2008.

\bibitem{DBLP:conf/nips/0002O14}
P.~Jain and S.~Oh.
\newblock Provable tensor factorization with missing data.
\newblock In {\em NIPS}, pages 1431--1439, 2014.

\bibitem{Koren2010}
Yehuda Koren.
\newblock Collaborative filtering with temporal dynamics.
\newblock {\em Commun. ACM}, 53(4):89--97, April 2010.

\bibitem{LeeH2014}
Dokyun Lee and Kartik Hosanagar.
\newblock Impact of recommender systems on sales volume and diversity.
\newblock In {\em Proc. Int. Conf. Inf. Syst.}, Auckland, New Zealand, December
  2014.

\bibitem{DBLP:conf/ijcai/LiZLZXW11}
Bin Li, Xingquan Zhu, Ruijiang Li, Chengqi Zhang, Xiangyang Xue, and Xindong
  Wu.
\newblock Cross-domain collaborative filtering over time.
\newblock In {\em IJCAI}, pages 2293--2298, 2011.

\bibitem{DBLP:conf/icdm/LiuDLLY03}
Bing Liu, Yang Dai, Xiaoli Li, Wee~Sun Lee, and Philip~S. Yu.
\newblock Building text classifiers using positive and unlabeled examples.
\newblock In {\em ICML}, pages 179--188, 2003.

\bibitem{DBLP:journals/pami/LiuMWY13}
Ji~Liu, Przemyslaw Musialski, Peter Wonka, and Jieping Ye.
\newblock Tensor completion for estimating missing values in visual data.
\newblock {\em {IEEE} Trans. Pattern Anal. Mach. Intell.}, 35(1):208--220,
  2013.

\bibitem{DBLP:conf/pkdd/NaritaHTK11}
Atsuhiro Narita, Kohei Hayashi, Ryota Tomioka, and Hisashi Kashima.
\newblock Tensor factorization using auxiliary information.
\newblock In {\em ECML/PKDD}, pages 501--516, 2011.

\bibitem{DBLP:conf/uai/RendleFGS09}
Steffen Rendle, Christoph Freudenthaler, Zeno Gantner, and Lars
  Schmidt{-}Thieme.
\newblock {BPR:} bayesian personalized ranking from implicit feedback.
\newblock In {\em UAI}, pages 452--461, 2009.

\bibitem{DBLP:conf/icml/RennieS05}
Jason D.~M. Rennie and Nathan Srebro.
\newblock Fast maximum margin matrix factorization for collaborative
  prediction.
\newblock In {\em ICML}, pages 713--719, 2005.

\bibitem{DBLP:conf/icml/SalakhutdinovM08a}
Ruslan Salakhutdinov and Andriy Mnih.
\newblock Bayesian probabilistic matrix factorization using markov chain monte
  carlo.
\newblock In {\em ICML}, pages 880--887, 2008.

\bibitem{scott2012calibrated}
Clayton Scott et~al.
\newblock Calibrated asymmetric surrogate losses.
\newblock {\em Electronic Journal of Statistics}, 6:958--992, 2012.

\bibitem{Sexton2013}
Robert~L. Sexton.
\newblock {\em Exploring Economics}.
\newblock Cengage Learning, Boston, MA, 2013.

\bibitem{Steiner1976}
Robert~L. Steiner.
\newblock The prejudice against marketing.
\newblock {\em J. Marketing}, 40(3):2--9, July 1976.

\bibitem{SunPV2014}
John~Z. Sun, Dhruv Parthasarathy, and Kush~R. Varshney.
\newblock Collaborative {K}alman filtering for dynamic matrix factorization.
\newblock {\em {IEEE} Trans. Signal Process.}, 62(14):3499--3509, 15 July 2014.

\bibitem{DBLP:conf/kdd/WangCGDKCMS15}
Yichen Wang, Robert Chen, Joydeep Ghosh, Joshua~C. Denny, Abel~N. Kho, You
  Chen, Bradley~A. Malin, and Jimeng Sun.
\newblock Rubik: Knowledge guided tensor factorization and completion for
  health data analytics.
\newblock In {\em SIGKDD}, pages 1265--1274, 2015.

\bibitem{DBLP:conf/sdm/XiongCHSC10}
Liang X., Xi~C., Tzu{-}Kuo H., Jeff~G. S., and Jaime~G. C.
\newblock Temporal collaborative filtering with bayesian probabilistic tensor
  factorization.
\newblock In {\em SDM}, pages 211--222, 2010.

\bibitem{DBLP:conf/hcomp/YiJJJ13}
Jinfeng Yi, Rong Jin, Shaili Jain, and Anil~K. Jain.
\newblock Inferring users' preferences from crowdsourced pairwise comparisons:
  {A} matrix completion approach.
\newblock In {\em First {AAAI} Conference on Human Computation and
  Crowdsourcing (HCOMP)}, 2013.

\end{thebibliography}
\end{document}